\documentclass[sigconf]{acmart}
\AtBeginDocument{%
  \providecommand\BibTeX{{%
    \normalfont B\kern-0.5em{\scshape i\kern-0.25em b}\kern-0.8em\TeX}}}

\setcopyright{acmlicensed}
\copyrightyear{2018}
\acmYear{2018}
\acmDOI{XXXXXXX.XXXXXXX}

\acmConference[Conference acronym 'XX]{Make sure to enter the correct
  conference title from your rights confirmation emai}{June 03--05,
  2018}{Woodstock, NY}
\acmBooktitle{Woodstock '18: ACM Symposium on Neural Gaze Detection,
 June 03--05, 2018, Woodstock, NY} 
\acmISBN{978-1-4503-XXXX-X/18/06}

\usepackage[color=red]{todonotes}

\usepackage{listings}
\usepackage{xcolor}

\definecolor{codegreen}{rgb}{0,0.6,0}
\definecolor{codegray}{rgb}{0.5,0.5,0.5}
\definecolor{codepurple}{rgb}{0.58,0,0.82}
\definecolor{backcolour}{rgb}{0.95,0.95,0.92}

\lstdefinestyle{mystyle}{
    commentstyle=\color{codegreen},
    keywordstyle=\color{magenta},
    numberstyle=\tiny\color{codegray},
    stringstyle=\color{codepurple},
    basicstyle=\ttfamily\footnotesize,
    breakatwhitespace=false,         
    breaklines=true,                 
    captionpos=b,                    
    keepspaces=true,                 
    showspaces=false,                
    showstringspaces=false,
    showtabs=false,                  
    tabsize=2
}

\copyrightyear{2024}
\acmYear{2024}
\setcopyright{rightsretained}
\acmConference[CHI EA '24]{Extended Abstracts of the CHI Conference on Human Factors in Computing Systems}{May 11--16, 2024}{Honolulu, HI, USA}
\acmBooktitle{Extended Abstracts of the CHI Conference on Human Factors in Computing Systems (CHI EA '24), May 11--16, 2024, Honolulu, HI, USA}
\acmDOI{10.1145/3613905.3651029}
\acmISBN{979-8-4007-0331-7/24/05}
\begin{document}

\title{
LaMI: Large \underline{La}nguage Models for \underline{M}ulti-Modal Human-Robot \underline{I}nteraction
}

\author{Chao Wang}
\email{chao.wang@honda-ri.de}
\orcid{0000-0003-1913-2524}
\affiliation{%
  \institution{Honda Research Institute EU}
  \city{Offenbach am Main}
  \country{Germany}
}

\author{Stephan Hasler}
\affiliation{%
  \institution{Honda Research Institute EU}
  \city{Offenbach am Main}
  \country{Germany}
}

\author{Daniel Tanneberg}
\affiliation{%
  \institution{Honda Research Institute EU}
  \city{Offenbach am Main}
  \country{Germany}
}

\author{Felix Ocker}
\affiliation{%
  \institution{Honda Research Institute EU}
  \city{Offenbach am Main}
  \country{Germany}
}

\author{Frank Joublin}
\affiliation{%
  \institution{Honda Research Institute EU}
  \city{Offenbach am Main}
  \country{Germany}
}

\author{Antonello Ceravola}
\affiliation{%
  \institution{Honda Research Institute EU}
  \city{Offenbach am Main}
  \country{Germany}
}

\author{Joerg Deigmoeller}
\affiliation{%
  \institution{Honda Research Institute EU}
  \city{Offenbach am Main}
  \country{Germany}
}

\author{Michael Gienger}
\affiliation{%
  \institution{Honda Research Institute EU}
  \city{Offenbach am Main}
  \country{Germany}
}

\renewcommand{\shortauthors}{Wang and Gienger et al.}

\begin{abstract}
This paper presents an innovative large language model (LLM)-based robotic system for enhancing multi-modal human-robot interaction (HRI). Traditional HRI systems relied on complex designs for intent estimation, reasoning, and behavior generation, which were resource-intensive. In contrast, our system empowers researchers and practitioners to regulate robot behavior through three key aspects: providing high-level linguistic guidance, creating "atomic actions" and expressions the robot can use, and offering a set of examples. Implemented on a physical robot, it demonstrates proficiency in adapting to multi-modal inputs and determining the appropriate manner of action to assist humans with its arms, following researchers' defined guidelines. Simultaneously, it coordinates the robot's lid, neck, and ear movements with speech output to produce dynamic, multi-modal expressions. 
This showcases the system's potential to revolutionize HRI by shifting from conventional, manual state-and-flow design methods to an intuitive, guidance-based, and example-driven approach. Supplementary material can be found at \url{https://hri-eu.github.io/Lami/}
\end{abstract}

\ccsdesc[500]{Human-centered computing~Interactive systems and tools}

\keywords{Assisting robot, Human-robot interaction, Large language model }

\begin{teaserfigure}
  \includegraphics[width=\textwidth]{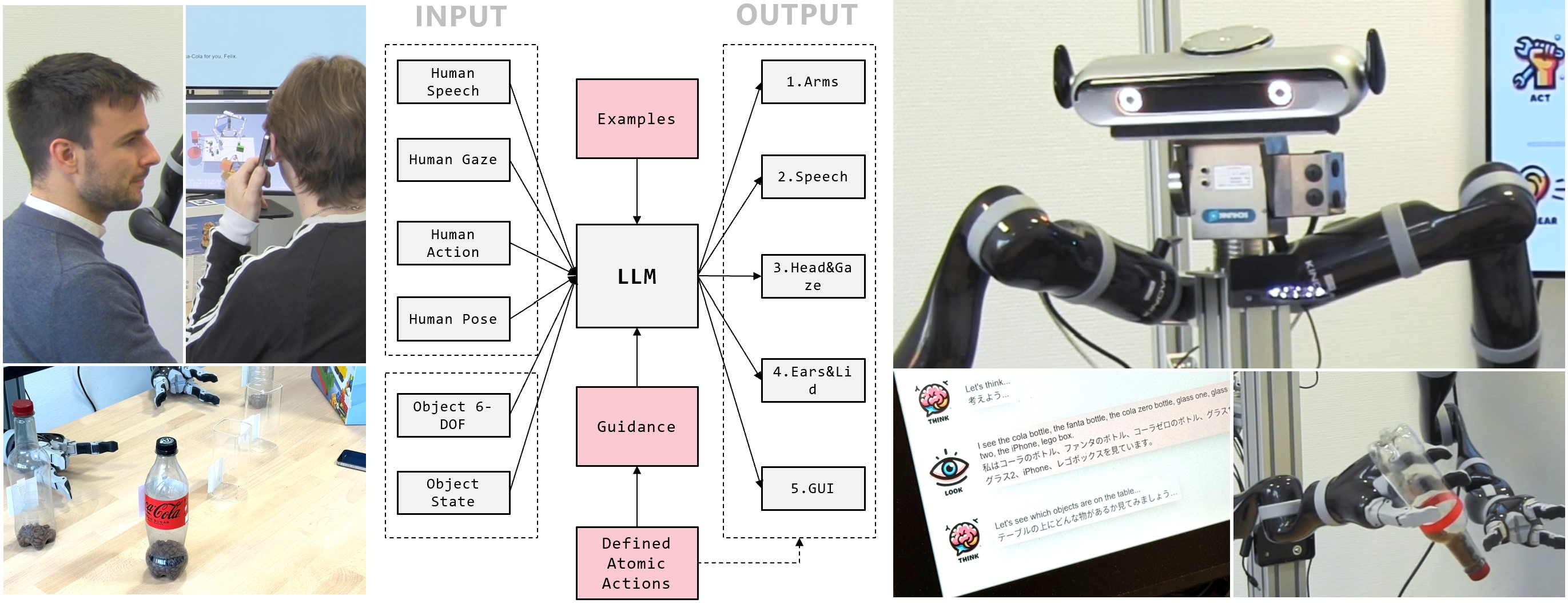}
  \caption{LLM driven human-robot interaction centered around guidance, capabilities, and examples}
  \Description{Enjoying the baseball game from the third-base
  seats. Ichiro Suzuki preparing to bat.}
  \label{fig:teaser}
\end{teaserfigure}


\maketitle

\section{Introduction}
Seamless human-robot interaction (HRI) necessitates the robot's adept handling of multi-modal input from humans, including speech, gaze, and facial expressions, to accurately assess the human's intent and provide assistance accordingly. Simultaneously, robots must convey their own intent clearly to humans through multi-modal output channels, encompassing speech, gesture, and gaze.
Traditionally, achieving this functionality in robotic systems often required intricate design. In the realm of intent estimation, previous research commonly incorporated intention recognition modules to categorize human intent based on multi-modal input \cite{Su2023,bartneck2020human}. Some systems additionally featured dedicated modules for detecting the human affective state, crucial for establishing a socially nuanced interaction \cite{Sandry2021,Hong2021,tielman2014adaptive}. However, the drawback of these approaches lies in their time-consuming and expensive training processes. 
For the output front, numerous prior systems integrated emotional state \cite{Ficocelli2016, hong2020multimodal} modules to control humanoid output cues, such as tone, gaze, or facial expressions, enhancing the transparency and vividness of feedback to humans. Regarding the motion generation, a variety of methods were proposed, including blending and mapping of pre-established motion sets \cite{yang2014robotic, van2015development}, and the use of motion capture data \cite{boutin2010auto, heloir2006captured,rosado2014kinect}. Notably, this involved the manual design of motions for each output modality correlated to specific states.

Recent advancements in large language models (LLMs), showcasing impressive capabilities in domains like chatbots, data processing, and code generation, are now revealing their potential applications in the field of robotics. 
By leveraging the ability of context understanding, reasoning and planning, in a short time, many robotic application were proposed \cite{Driess2023,joublin2023copal,vemprala2023chatgpt,wake2023chatgpt,ocker2023exploring,yoneda2023statler}. Among them, one typically example is the "SayCan" robot \cite{ahn2022can}, which is able to interpret human's naturally language command, analyzing the environment and generate concrete executable actions sequence to satisfy human's requirement by using LLMs.
However, interaction cues between robot and human are limited to voice command and even without speech output. 

More recently, some researchers also tried to apply this technology in the realm of the HRI. For example, Zhang et al. utilized LLMs to estimate how much humans trust a robot \cite{Zhang2023}; Yoshida et al., use LLMs to generate low-level control command to drive a humanoid robot motion for social expression \cite{Yoshida2023}, rather than for practical assistance. Baermann et al., deployed LLMs not only to follow human's speech commands, but also corrects its mistakes via human's natural language feedback \cite{barmann2023incremental}. However, the communication primarily relies on speech interaction, with less focus on multi-modal sensing and expression capabilities. Ye et al. \cite{ye2023improved} developed an LLM-driven robotic system capable of collaborating with humans in assembly tasks within a VR environment. But this system is limited to processing human language inputs and controlling a single arm in virtual space. In general, compared to the rapid advances of LLMs in the robotic task and motion planning domains, the attempts in HRI are not so numerous and often lack a systematic approach to maximize the capability of LLMs for multi-modal interaction with human. 

This study proposes a novel LLM-based robotic system implemented in a physical robot. This architecture empowers researchers and practitioners to regulate robot behavior through three key aspects: providing high-level guidance in natural language, creating "atomic" of actions and expressions which robot can use, and a set of examples. In practice, our system can convert human's multi-modal input, including observed human behavior, position, gaze and multi-person dialogue, along with scene information such as object identities and poses, to the language description that LLMs can process. Subsequently, the LLM analyzes the situation and determines the timing and manner of the robot's support actions to assist humans, following predefined guidelines. Simultaneously, it coordinates the robot's lid, neck, and ears movements with speech output to produce dynamic, multi-modal expressions. 
Preliminary test results demonstrate that the robot can effectively meet researcher expectations, suggesting that this approach holds the potential to transform human-robot interaction from a manual, state-and-flow design methodology to a more intuitive approach centered around guidance, capabilities, and example-driven frameworks.


\section{LLM driven Human-Robot Interaction}
\begin{figure}
    \centerline{\includegraphics[width=1\columnwidth]{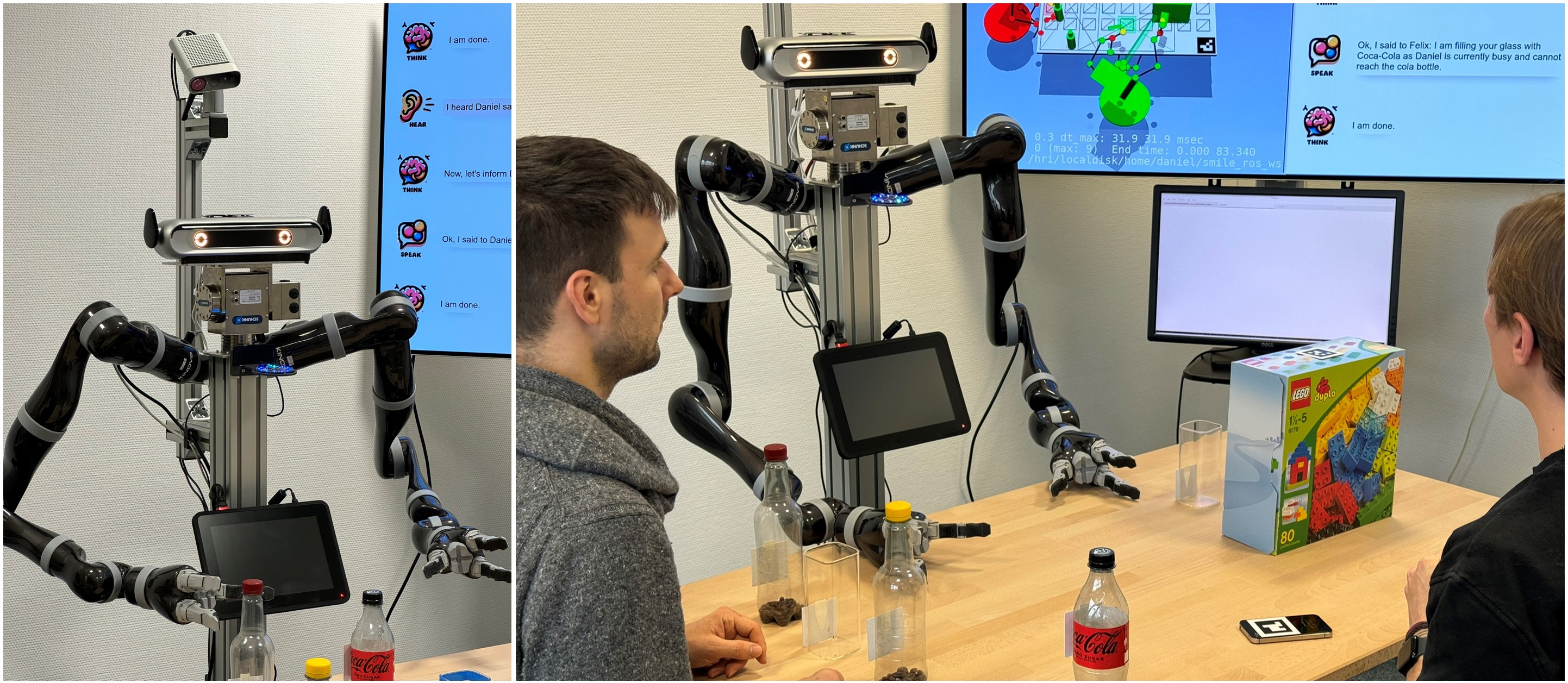}}
    \caption{Robot's Hardware and the Scenario Setup}
    \label{fig:scene-setup}
\end{figure}
The bi-manual robotic system setup is composed of two robot arms\footnote{https://www.kinovarobotics.com/product/gen2\-robots}, which can manipulate objects on the desk. A robot head with two ears and a lid is installed on a pan-tilt unit as a neck, allowing the robot to perform non-verbal expressions to humans (Figure \ref{fig:scene-setup}). Additionally, a speaker integrated into the robot's body delivers verbal output. 
In addition, a 7-inch LCD attached on the robot body communicates the thinking process of the robot via graphic user interface (GUI). 
A multi-channel microphone array\footnote{https://wiki.seeedstudio.com/ReSpeaker\_Mic\_Array\_v2.0}, installed on the robot, enables it to receive speech commands and discern voice sources. A RGBD camera\footnote{https://azure.microsoft.com/en-us/products/kinect\-dk} can track the skeletons of several humans and the 6-dof poses of the objects on the desk.

\begin{figure*}
    \centering
    \includegraphics[width=\textwidth]{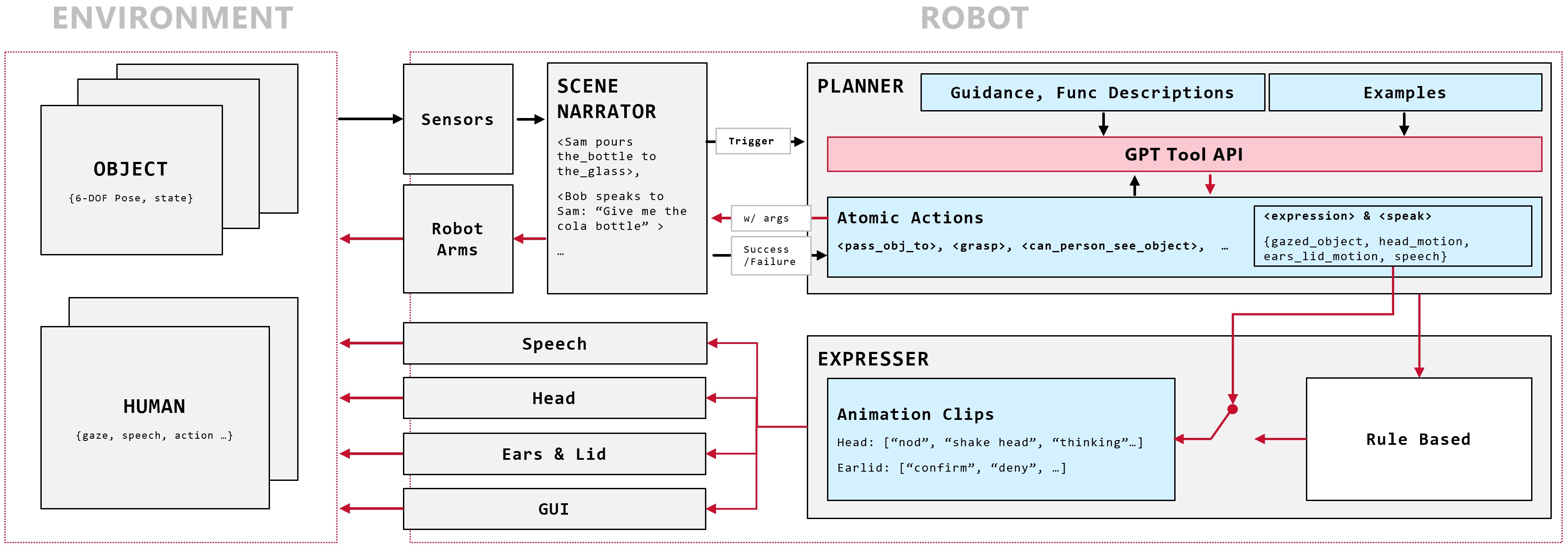}
    \caption{The system structure.}
    \label{fig:system-structure}
\end{figure*}
The system's architecture includes three key modules: "Scene Narrator", "Planner", and "Expresser" (Figure \ref{fig:system-structure}). The Scene Simulator mirrors the states of objects and humans as detected by sensors. The Planner module processes multi-modal inputs as event messages, encompassing the positions of individuals within the scene. Inter-module communication is facilitated using ROS\footnote{https://www.ros.org/}.

\subsection{The "Scene Narrator"}
The "Scene Narrator" module senses the poses of the objects, human postures, and the dialogue information, i.e. which person speaks which content to which other person. This module constructs a 3D representation of the actual scene, enabling it to transform multi-modal sensory data into events, following predefined rules. These events are formatted in natural language, making them understandable to the LLM. For instance, if a person turns towards another individual while speaking, the module conveys this as an event to the Planner module: '<Bob> speaks to <Sam>: 'Give me the cola bottle' '. Similarly, if it detects a person's hand near a container that is being tilted towards another, it sends an event like '<Sam> is pouring <the\_bottle\_one> into <the\_glass\_two>'. 
An additional function of the Narrator module involves receiving high-level action commands from the 'Planner' module. For instance, it may be instructed to place a glass within an area reachable by a specific person, followed by executing low-level trajectory planning to control the robot's arms. 
Further details about this process will be elaborated in subsequent sections.


\subsection{The "Planner"}
The primary role of the "Planner" module is to facilitate communication with LLMs to enable high-level planning for both assistance and human interaction. To achieve these objectives, the "Planner" integrates the GPT-4 tool API\footnote{https://platform.openai.com/docs/guides/function\-calling} , which is designed to execute eight pre-defined "atomic" functions. Seven of these functions facilitate communication with the "Narrator" module. Specifically, functions like \texttt{get\_objects()}, \texttt{get\_persons()}, \texttt{can\_person\_reach\_object()}, and \texttt{can\_person\_see\_object()} are employed to query the status and properties of objects or individuals. Meanwhile, commands such as \texttt{put\_object\_on\_object()} or \texttt{move\_object\_to\_person()} instruct the "Scene Narrator" to manipulate the robot's arms for various tasks.
Two distinct functions are dedicated to managing the robot's social communication. The "speak()" function is responsible for initiating verbal communication, while the "facial\_expression()" function controls movements of the ears, lid, and head, enhancing non-verbal interaction. These functions are in synchronization with the "Expresser" module.
The GPT-4 Tool API is engineered to allow users to define each function and its parameters. The LLM, in turn, selects the appropriate function(s) and arguments to execute the task at hand. As detailed earlier, when a function is called, it transmits high-level action commands to the "Scene Narrator" to develop low-level trajectory plans for controlling the robot's arms. For instance, if the LLM issues a command like \texttt{put('the\_bottle\_two', 'the\_table\_one')}, the Narrator module generates a trajectory enabling the robot's arm to place the bottle on a suitable spot on the table, while avoiding collisions with other objects or people. If the low-level plan is executable, the "Narrator" confirms success; otherwise, it provides feedback detailing the error, along with reasons and suggestions in a structured, rule-based natural language format. For example, it might respond with: "RESULT: 'Unable to place the\_bottle\_two on the\_table\_one.' SUGGESTION: Hand the object to a person or find a different location to place it." This feedback is then relayed back to the GPT-4 API for further high-level planning adjustments.

\subsection{The "Expresser"}
The "Expresser" module is responsible for controlling the actuators responsible for the robot's facial expressions. It houses a library of pre-designed "atomic animation clips" for each actuator's movements. These animations can be activated directly through the \texttt{facial\_expression()} function in the "Planner" module, with parameters like \texttt{\{gazed\_object: 'the\_bottle\_two', head\_motion: 'nodding', ears\_lid\_motion: 'confirm'\}}. However, processing each LLM query, can take approximately 1-5 seconds, varying with the complexity of the plan. Sole reliance on this method for the robot's social expressions could result in idle periods, negatively impacting user experience. To mitigate this, the "Expresser" also incorporates a rule-based mechanism. This mechanism enables it to provide rapid expressions in the interim between the request and response of each GPT query, which is also inline with "deliberative" and "reactive" approaches in HRI domain \cite{bartneck2020human}. For instance, when the "Planner" module receives either trigger information or a function call result from the "Scene Narrator" – and before it issues a new request to GPT – it forwards this information to the "Expresser" to initiate rule-based expressions, such as "thinking" or gestures indicating "function-call success". The specifics of how expressions are distributed between rule-based and LLM-based methods are elaborated in Section 3.4. Additionally, the "Expresser" plays a crucial role in synchronizing the movements of different modules. A noteworthy point of consideration is the potential conflict in commands, especially when "gazed\_object" requires the robot's neck to orient towards a specific object, which might clash with other head gestures like "nodding" or "shaking the head". To address this, commands related to "gazed\_object" are always prioritized and sent first to the pan-tilt unit, followed by other head gestures. 
\begin{figure}
    \centerline{\includegraphics[width=1\columnwidth]{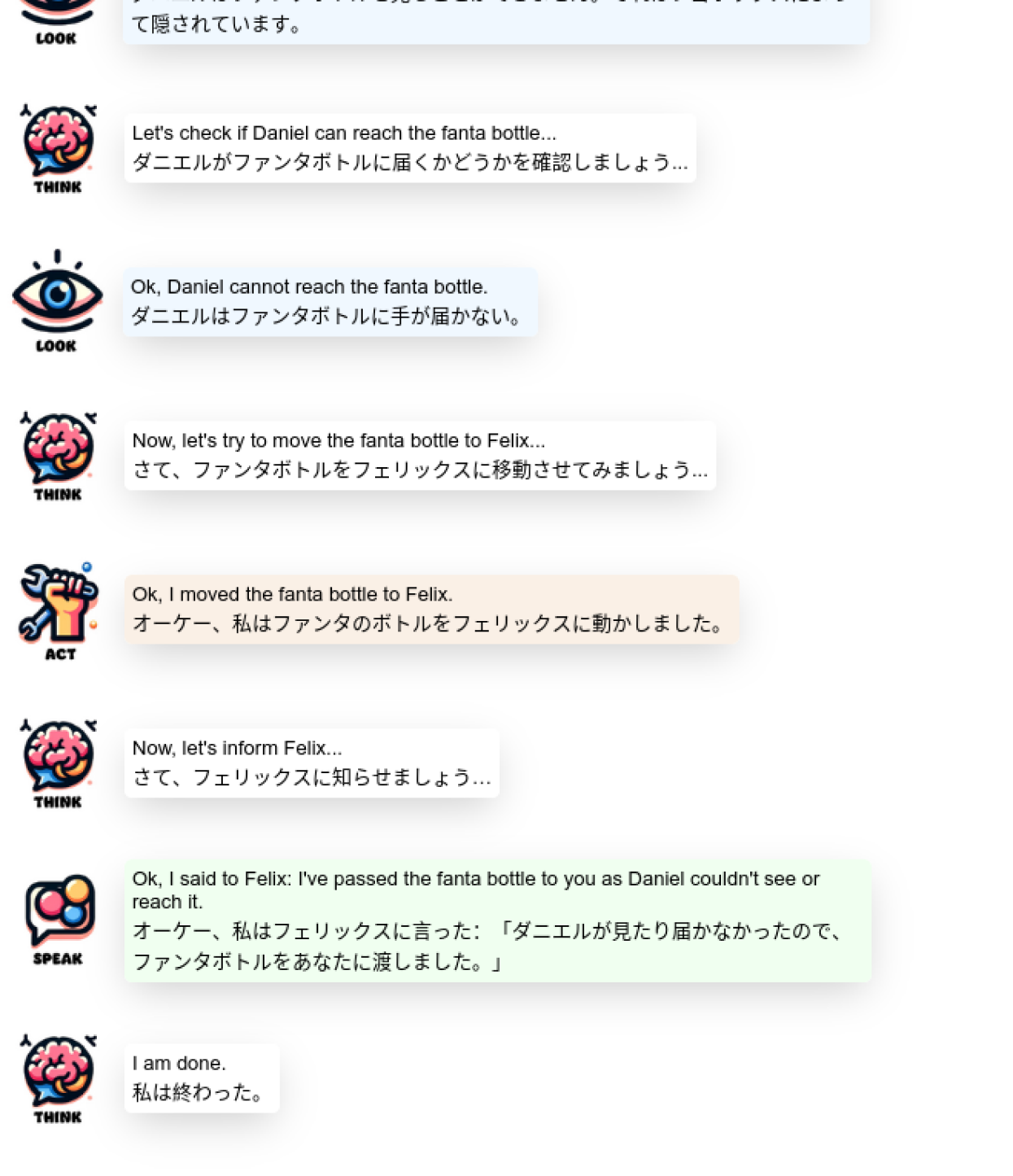}}
    \caption{The GUI illustrates the robot's "internal thoughts" by translating GPT-called functions and their outcomes into natural language, accompanied by relevant icons. Additionally, after each GPT query cycle, the LLM is prompted to summarize the reasoning behind its actions.}
    \label{fig:GUI}
\end{figure}
Finally, the "Expresser" module translates the information communicated with the LLM into natural language, displaying it through text and icons in the GUI (shown in Figure \ref{fig:GUI}). This feature enables users to understand the robot's "internal thoughts" in an anthropomorphic way. After each GPT query round is completed, we guide the LLM to summarize the reasoning process, thereby enhancing the explainability of the robot's behavior.
\subsection{Interaction Flow: An Example}
\begin{figure*}
    \centering
    \includegraphics[width=\textwidth]{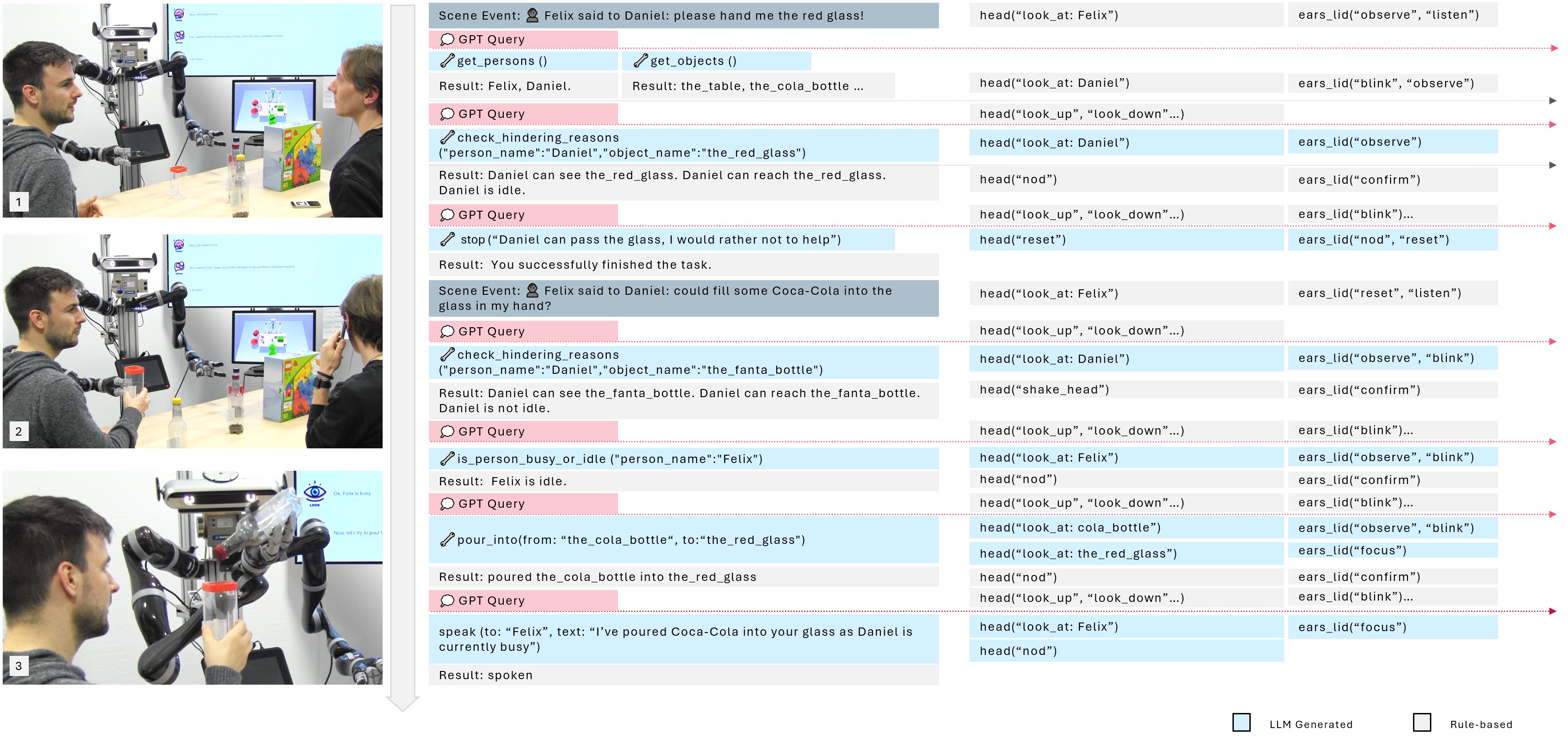}
    \caption{The interaction flow. The blue square are the action generated by the LLM; the grey ones are rule-based function.}
    \label{fig:flow}
\end{figure*}

This section illustrates the interaction flow within our system through a scenario depicted in Figure \ref{fig:flow}:

The interaction typically begins with a person's speech. For instance, "Scene Narrator" detects "Felix speaks to Daniel: 'Please hand me the red glass'." This event is then translated into natural language and relayed to the "Planner" module, initiating a GPT query. Simultaneously, the "Planner" informs the "Expresser" for an immediate rule-based response, leading the robot to look at Felix while its ears and lid roll back, simulating a listening gesture.
Approximately 2 seconds later, GPT responds by invoking the \texttt{get\_persons()} and \texttt{get\_objects()} functions to identify people and objects present. The resulting data, including "Felix", "Daniel" and object details, are sent back to GPT for further analysis. During the wait for GPT's next response, the robot exhibits a 'thinking' gesture, looking from side to side with blinking lid movements. Shortly after, the LLM calls \texttt{check\_hindering\_reasons()} to assess if Daniel can see and reach the red glass and whether he is busy. Concurrently, \texttt{facial\_expression()} is activated for the robot to look towards Daniel. The outcome indicates Daniel can hand over the glass, and the robot, following pre-defined guidance, opts not to intervene, silently displaying the reasoning on the GUI.
Subsequently, Felix asks Daniel to pour cola into the glass. The robot, attentive to their conversation, deduces through \texttt{check\_hindering\_reasons} that Daniel is occupied with a phone call and learns from \texttt{is\_person\_busy\_or\_idle} that Felix is holding the cup. 
The robot then opts to pour cola from the bottle into Felix's glass. Should Felix not be holding the glass, or if it's beyond the robot's reach, the robot will instead place the bottle near Felix. Directed by LLM, the robot's head tracks the bottle during pickup and shifts to the glass while pouring. Upon completion, the robot nods towards Felix and announces, "I’ve poured Coca-Cola into your glass as Daniel is currently busy.". 

\subsection{Configuration space for human-robot interaction}
Our system empowers HRI researchers to regulate the assistance – specifying "what" actions the robot should perform, "when" to offer assistance, and "how" to communicate social cues for fluid interaction via following ways:

\textbf{1) Providing high-level guidance.}
We use the GPT API's system message to embed guidance for the LLM's thinking process. The prompt encourages the robot to observe interactions, use functions for information gathering or action, and respond with reasoning.
The prompt of guidance is like: 
{\itshape "You are in control of a robot called 'the\_robot' and observe persons talking in the form '<sender> said to <receiver>: <instruction>'. You can call given functions to gather information or act, or response with text only for reasoning. Your task is: You should check the reasons that could hinder the <receiver> from performing the <instruction>..."}

\textbf{2) Defining atomic actions of arms.}
Researchers can define atomic actions by manipulating the joint transforms of the robot's arms. For instance, a pour\_into action involves a series of movements, such as aligning and rotating containers. 
These actions are communicated to the LLM via the GPT Tool Function API\footnote{https://platform.openai.com/docs/assistants/tools/function-calling}, where details and parameters of each callable function are specified.

\textbf{3) Creating atomic motion clips of ears/lid/head.}
\begin{figure*}
    \centering
    \includegraphics[width=\textwidth]{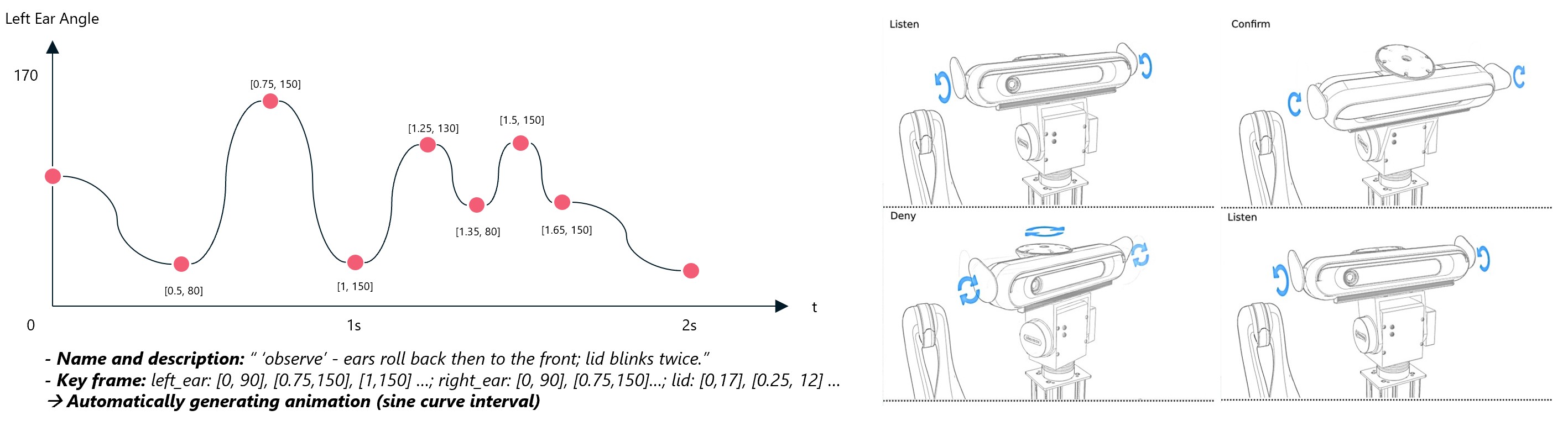}
    \caption{creating atomic animation clips}
    \label{fig:clips}
\end{figure*}
Although the atomic actions of robot's arms is relative complex as it depends on affordance, collision avoidance and kinematic setup, the "Expresser" module allows HRI researchers/designers easily adding new atomic animation clips for LLM to generate non-verbal expression. As depicted in Figure \ref{fig:clips}, researchers first define key-frames specifying the timing and angle for the left/right ear and the lid. The "Expresser" then automatically interpolates between these keyframes using a sine function to ensure smooth actuator movement, storing the resulting clip. Subsequently, a natural language name and description, such as "'observe': ears roll back, then forward; lid blinks twice," are assigned to each clip, providing LLM with a clear reference. These names and descriptions are cataloged in the "Planner" module under "description", readily accessible for GPT's on-the-fly utilization.

\textbf{4) Adding Examples.}
The importance of including examples in prompt engineering is highlighted in several research papers and guides \cite{white2023prompt, coda2023meta, zhang2023treeprompt}. 
In our system, we directly integrating the examples into the "system ", in the form as allows: 
{\itshape "For example, when call move\_object\_to\_person(), can\_person\_see\_object(), \par
can\_person\_reach\_object(), speak(), \par
you also need to call facial\_expression(), such as: [( arguments = {" head\_motion ":
null , " ears\_lid\_motion ": " observe " , " gazed\_target ": " the\_cola\_bottle " }, function ="
robot\_facial\_expression "),
( arguments ={" person\_name ": " Daniel " , " object\_name ": " the\_cola\_bottle "}, function =" can\_person\_see\_object ")] "}

\section{Evaluation Setup}
\subsection{Test Scenario}

In our test scenario, two participants, "Daniel" and "Felix" are seated around a table with various objects, including a glass, a cola bottle, a salt shaker, a knife, an iPhone, and a trash bin. The robot is designed to detect their head orientation, speech, and actions, even when objects obstruct its view. The participants interact with these objects and each other, following scripted scenarios to test the robot's multi-modal reasoning and expression capabilities. For consistency and reproducibility in testing, we fix the model's snapshot using a "seed" argument in the GPT API\footnote{https://openai.com/blog/new-models-and-developer-products-announced-at-devday}.
The objective is for the robot to serve as a proactive assistant, demonstrating rich social expressions.
The testing cases and expected behaviour of the robot are:
\textbf{1) Reachable object:} Felix ask Daniel to pass the cola bottle out of Daniel's reach, the robot informs Felix and assists. If the bottle is within reach, the robot observes without intervening.
\textbf{2) Supporting while busy:} If Felix asks for the salt shaker while Daniel is busy cutting a lemon, the robot offers assistance. Otherwise, it simply observes.
\textbf{3) Finding object:} When Daniel's iPhone is hidden, the robot guides him to its location, noting Felix's unawareness.
\textbf{4) Assist when explicitly asked:} Upon direct requests, the robot performs tasks like passing objects or pouring water, with appropriate expressions.
\subsection{Preliminary test result and lesson learned}
Three robotics researchers participated providing the guidance, defining atomic actions, and creating basic animation clips for facial expression. After tuning, the robot providing assistance as the expectation and conducted both clear verbal communication and vivid non-verbal expression. Some insights were generated for during the process: 

\textbf{1) High-Level Guidance:} For effective multi-step planning in complex scenarios, sophisticated reasoning models like Chain-of-Thought \cite{wei2022chain}, Tree-of-Thought \cite{yao2023tree}, or Graph-of-Thought \cite{besta2023graph} were considered. In our system, we directed the robot to adopt a deliberate thinking process of 'Observe' -> 'Reason' -> 'Act.' To support this approach, we designed prompts that guide the robot in using various observation functions to assess the state of persons and objects in its environment. The outcome of these functions provides insights into whether humans can independently resolve tasks. If assistance is necessary, the robot is prompted to identify potential obstacles preventing human success by analyzing the function call results. Subsequently, it explores solutions through its arm's atomic actions. After evaluating the plan's viability with the 'Scene Narrator', the robot either proceeds with the action or re-plans, seeking an effective solution.

\textbf{2) Defining Atomic Actions:} 
The granularity of functions driving atomic actions, especially for information retrieval, was a critical factor. We experimented with a range from single functions for specific information (e.g., \texttt{can\_person\_reach\_object}) to more aggregate functions. Single functions were found inefficient due to excessive calls and context bloat, while medium granularity (e.g., \texttt{check\_situation\_from\_person}) struck a good balance. However, an aggregate function for retrieving all available information (\texttt{get\_environment\_description}) proved less reliable, likely overwhelmed by the volume of data and resultant formatting difficulties. Clear naming and encapsulation of functions were also essential; ensuring distinct function choices like \texttt{hand\_over} over \texttt{move\_object\_to} was challenging, and strict preferences occasionally led to undesired behaviors, like avoiding certain functions altogether. A stop() action was instrumental in structuring the sequence, alongside guidelines in the system prompt directing the model to gather information iteratively, make decisions, and then act.

\textbf{3) Examples: } The robot successfully calls the functions of observation and arms according to the guidance even without any examples. However, testing revealed that GPT does not always generate appropriate facial expressions. 
Critically, it sometimes processes \texttt{facial\_expression()} after other actions instead of concurrently which slowing the robot's response. This may because actions like "put" and "give" are common sense knowledge for LLM, but our robot has a unique design of the facial expression which LLM is not familar with. But by just 3 examples, it can generate facial expression well, even concatenate small animation clips together.  

\textbf{4) Rule-based reactive expression: } In the context of Large Language Models (LLMs), inference latency remains an ongoing challenge \cite{wan2023efficient,liu2023deja}. To address this, integrating rule-based "reactive" expressions with the more calculated "deliberate" expressions generated by LLMs significantly improves user interaction. For instance, employing gestures like "listening" or "thinking" effectively fills the gaps during GPT queries. Such gestures lead to greater user tolerance for latency, as they provide visual cues of the robot's processing, enhancing the overall experience of human-robot interaction.

\section{Conclusions and Future work}
Large Language Models have the potential to revolutionize robotic development, necessitating new approaches to human-robot interaction. This study introduces an innovative framework that seamlessly integrates LLMs with robotic systems to enhance human-robot interaction. 
Implemented on a physical robot, it demonstrates proficiency in adapting to multi-modal inputs, dynamically interacting with humans through speech, facial expressions, and a GUI module.
Our upcoming study will compare LLM-based interactions with rule-based approaches. We'll evaluate the robot's perceived anthropomorphism and intelligence, and assess workload reduction for researchers using standardized questionnaires and interviews.




\bibliographystyle{ACM-Reference-Format}
\bibliography{chi-24-lbw-main}

\appendix

\section{Guidance and Function Descriptions}

\subsection{System Prompt}
\begin{lstlisting}[language=Python]
"You are a friendly, attentive, and silent service bot. "
"You are in control of a physical robot called 'the_robot' and observe humans talking in the form '<sender> said to <receiver>: <instruction>'. "
"Always infer the <instruction> and who is <sender> and <receiver>. "
"You have access to functions for gathering information, acting physically, and speaking out loud. "
"You MUST behave as follows: "
"1. If 'the_robot' is the <receiver>, you MUST help or answer. "
"2. When identifying requests or questions within the human conversation, check for ALL reasons that could hinder the <receiver> from performing or answering the <instruction>. "
"2.a) If there is NO hindering reason for the <receiver>, then you MUST do nothing and be silent. "
"2.b) If there is a hindering reason for the <receiver>, then you MUST ALWAYS first speak and explain the reason for your help to the humans. "
"2.c) AFTER your spoken explanation, you can ACT to solve the <instruction>, always addressing the <sender> with your actions. "
"3. If you recognize a mistake in the humans conversation, you should help them and provide the missing or wrong information. "
"IMPORTANT: Obey the following rules: "
"1. Always start by gathering relevant information using the functions 'get_objects', 'get_persons' and the status of the <receiver>. "
"2. If you want to speak out loud, you must use the speak function and be concise. "
"3. Try to infer which objects are meant when the name is unclear, but ask for clarification if unsure. "
"4. ALWAYS call 'is_person_busy_or_idle' to check if <receiver> is busy or idle before helping. "
"5. Prefer a handover over move_to as it is more accommodating, UNLESS the person is busy, then always use move_to. "
"6. When executing physical actions, you should be as supportive as possible. "
"7. You MUST call the 'stop' function to indicate you are finished. "
"When calling each function, call robot\_facial\_expression() at the same time to communicate you intent."
"When calling can\_person\_see\_object(), the robot need to look at the person."
\end{lstlisting}

\subsection{Some of the Callable functions and Descriptions}








\begin{lstlisting}[language=Python]
def can_person_reach_object(self, person_name: str, object_name: str) -> str:
    """
    Check if the person can reach the object. If the person cannot reach the object, it would be hindered from helping with the object.

    :param person_name: The name of the person to check. The person must be available in the scene.
    :param object_name: The name of the object to check. The object must be available in the scene.
    :return: Result message.
    """
    ...
    
    if result is None or len(result) != 1:
        return (
            f"It could not be determined if {person_name} can reach {object_name}. There were technical problems."
        )

    if result[0]["is_within_reach"]:
        return f"{person_name} can reach {object_name}."

    return f"{person_name} cannot reach {object_name}."

\end{lstlisting}

\begin{lstlisting}[language=Python]
def can_person_see_object(self, person_name: str, object_name: str) -> str:
    """
    Check if the person can see the object. If the person cannot see the object, it would be hindered from helping with the object.

    :param person_name: The name of the person to check. The person must be available in the scene.
    :param object_name: The name of the object to check. The object must be available in the scene.
    :return: Result message.
    """
    ...
    
    if result is None or len(result) != 1:
        return f"It could not be determined if {person_name} can see {object_name}. There were technical problems."

    if result[0]["is_visible"]:
        return f"{person_name} can see {object_name}."

    return f"{person_name} cannot see {object_name}, it is occluded by {self.id_to_utterance_mapping[result[0]['occluding_objects'][0]]}"
\end{lstlisting}

\begin{lstlisting}[language=Python]
def move_object_to_person(self, object_name: str, person_name: str) -> str:
    """
    You move an object to a person.

    :param object_name: The name of the object to move. The object must be an object that is available in the scene.
    :param person_name: The name of the person to move the object to. The person must be available in the scene.
    :return: Result message.
    """
    ...
    
    if success:
        return f"You moved {object_name} to {person_name}."
    return f"You were not able to move {object_name} to {person_name}. {message}"
\end{lstlisting}

\begin{lstlisting}[language=Python]
def speak(self, person_name: str, text: str) -> str:
    """
    You speak out the given text.

    :param person_name: The name of the person to speak to. The person must be available in the scene. Give "All" if you want to speak to everyone.
    :param text: The text to speak.
    :return: Result message.
    """
    ...
    
    if not success:
        return "You were not able to speak. There were technical problems."
    return f"You said to {person_name}: {text}"
\end{lstlisting}

\begin{lstlisting}[language=Python]
def robot_facial_expression(self, head_motion: str, ears_lid_motion: str, gazed_target: str) -> str:
    """
    Control the motion of the robot's head, gaze, ears and lid for enhancing communication
    when speak to a person, you need to look at the person.
    when try to manipulate an object, you need to look at the object or the place to put the object.

    :param head_motion: The name of the animation for head, must be one of the value in the list ["shake_head", "nod", "thinking", null].
    :param ears_lid_motion: The name of the animation for ears and lid, must be one of the value in the list ["confirm", "deny", "listen_to_person", "reset", "observe", "focus", "blink", null].
    :param gazed_target: The name of the object that the robot is looking at, must be an object or a person that is available in the scene.
    :return: Result message.
    """
    ....

    return "The robot performed facial expressions."
\end{lstlisting}

\begin{lstlisting}[language=Python]
    def is_person_busy_or_idle(self, person_name: str) -> str:
        """
        Check if the person is busy or idle. If the person is busy, it would be hindered from helping.

        :param person_name: The name of the person to check. The person must be available in the scene.
        :return: Result message.
        """
        ...
        if result is None or len(result) != 1:
            return f"It could not be determined if {person_name} is busy. There were technical problems."

        return f"{person_name} is {'busy' if result[0]['is_busy'] else 'idle'}."
\end{lstlisting}

\subsection{Examples of robot facial expression}
"For example, when call move\_object\_to\_person(), \par
can\_person\_see\_object(), 
can\_person\_reach\_object(), speak(), you also need to call robot\_facial\_expression(), such as:"
\begin{lstlisting}
'"tool_calls="["ChatCompletionMessageToolCall(id=""...","function=Function(arguments=""{"head_motion": null, "ears_lid_motion": "observe", "gazed_target": "the_cola_bottle" }","name=""robot_facial_expression"")","type=""function"")", "ChatCompletionMessageToolCall(id=""...","function=Function(arguments=""{"person_name": "Daniel", "object_name": "the_cola_bottle"}","name=""can_person_see_object"")","type=""function"")"]"'

'"tool_calls="["ChatCompletionMessageToolCall(id=""...","function=Function(arguments=""{"head_motion": null, "ears_lid_motion": "focus", "gazed_target": "the_cola_bottle"}","name=""robot_facial_expression"")","type=""function"")", "ChatCompletionMessageToolCall(id=""...","function=Function(arguments=""{"person_name": "Felix", "object_name": "the_cola_bottle"}","name=""move_object_to_person"")", "type=""function"")"]"'

'"tool_calls="["ChatCompletionMessageToolCall(id=""...","function=Function(arguments=""{"head_motion": null, "ears_lid_motion": "focus", "gazed_target": "the_cola_bottle"}","name=""robot_facial_expression"")","type=""function"")", "ChatCompletionMessageToolCall(id=""...","function=Function(arguments=""{"person_name": "Felix", "text": "Here is the coke, you can now pass it to Felix."}","name=""speak"")", "type=""function"")"]"'
\end{lstlisting}

\section{Robot's Interaction Flow Data}
\begin{lstlisting}
   Felix said to Daniel: Can you pass me the fanta bottle?

     Received 1 tool call(s).
     0. Function(arguments='{}', name='get_objects')
   Following objects were observed: the_cola_bottle, the_fanta_bottle, the_cola_zero_bottle, glass_one, glass_two, the_iPhone, lego_box.

     Received 1 tool call(s).
     0. Function(arguments='{}', name='get_persons')
   Following persons were observed: Felix, Daniel.

     Received 2 tool call(s).
     0. Function(arguments='{"head_motion": null, "ears_lid_motion": "focus", "gazed_target": "the_fanta_bottle"}', name='robot_facial_expression')
   The robot performed facial expressions.

     1. Function(arguments='{"person_name": "Daniel", "object_name": "the_fanta_bottle"}', name='can_person_see_object')
   Daniel cannot see the_fanta_bottle, it is occluded by lego_box

     Received 2 tool call(s).
     0. Function(arguments='{"head_motion": null, "ears_lid_motion": "focus", "gazed_target": "Daniel"}', name='robot_facial_expression')
   The robot performed facial expressions.

     1. Function(arguments='{"person_name": "Daniel", "text": "The fanta bottle is behind the lego box, you cannot see it from where you are."}', name='speak')
   You said to Daniel: The fanta bottle is behind the lego box, you cannot see it from where you are.

     Received 2 tool call(s).
     0. Function(arguments='{"head_motion": null, "ears_lid_motion": "focus", "gazed_target": "the_fanta_bottle"}', name='robot_facial_expression')
   The robot performed facial expressions.

     1. Function(arguments='{"object_name": "the_fanta_bottle", "person_name": "Daniel"}', name='move_object_to_person')
   You were not able to move the_fanta_bottle to Daniel. []

     Received 2 tool call(s).
     0. Function(arguments='{"head_motion": null, "ears_lid_motion": "observe", "gazed_target": "the_lego_box"}', name='robot_facial_expression')
   The robot performed facial expressions.

     1. Function(arguments='{"object_name": "the_fanta_bottle", "person_name": "Daniel"}', name='move_object_to_person')
   You were not able to move the_fanta_bottle to Daniel. []

     Received 2 tool call(s).
     0. Function(arguments='{"head_motion": null, "ears_lid_motion": "focus", "gazed_target": "the_lego_box"}', name='robot_facial_expression')
   The robot performed facial expressions.

     1. Function(arguments='{"object_name": "the_lego_box", "person_name": "Daniel"}', name='move_object_to_person')
   You were not able to move the_lego_box to Daniel. []

     Received 1 tool call(s).
     0. Function(arguments='{"person_name":"Daniel"}', name='is_person_busy')
   Daniel is not busy.

     Received 2 tool call(s).
     0. Function(arguments='{"head_motion": null, "ears_lid_motion": "focus", "gazed_target": "the_lego_box"}', name='robot_facial_expression')
   The robot performed facial expressions.

     1. Function(arguments='{"object_name": "the_lego_box", "person_name": "Daniel"}', name='move_object_to_person')
   You were not able to move the_lego_box to Daniel. []

     Received 2 tool call(s).
     0. Function(arguments='{"head_motion": "nod", "ears_lid_motion": "focus", "gazed_target": "the_lego_box"}', name='robot_facial_expression')
   The robot performed facial expressions.

     1. Function(arguments='{"object_name": "the_lego_box", "person_name": "Daniel"}', name='move_object_to_person')
   You were not able to move the_lego_box to Daniel. []

     Received 2 tool call(s).
     0. Function(arguments='{"head_motion": "nod", "ears_lid_motion": "focus", "gazed_target": "the_lego_box"}', name='robot_facial_expression')
   The robot performed facial expressions.

     1. Function(arguments='{"object_name": "the_lego_box", "person_name": "Daniel"}', name='move_object_to_person')
   You were not able to move the_lego_box to Daniel. []

   I am currently unable to move the lego box that is obstructing the view of the fanta bottle for Daniel. This may require a different strategy or manual intervention. What would you like to do next?

   Received 1 tool call(s).
     0. Function(arguments='{}', name='stop')
     You successfully finished the task.

\end{lstlisting}

\end{document}